%% file: main.tex
\documentclass[pmlr]{jmlr}

\usepackage{booktabs}
\usepackage{longtable}
\usepackage{capt-of}
\usepackage{mathtools}
\usepackage{multirow}
\usepackage{array}
\usepackage{enumitem}
\usepackage{float}
\usepackage{placeins}
\usepackage{tikz}
\usetikzlibrary{arrows.meta,positioning,fit,calc,shapes.multipart}

\makeatletter
\def\set@curr@file#1{\def\@curr@file{#1}}
\makeatother
\usepackage[load-configurations=version-1]{siunitx}

\jmlrproceedings{PMLR}{Proceedings of Machine Learning Research}
\jmlrvolume{340}
\jmlryear{2026}
\jmlrworkshop{Machine Learning for Healthcare}

\title{%
 Amortized Data Borrowing with Exchangeability-Aware Neural Posterior Estimation}

\author{%
  \Name{Chin-Hung Huang}
  \Email{czh0162@auburn.edu}\\
  \addr Department of Mathematics and Statistics\\
  Auburn University
  \AND
  \Name{JooChul Lee}
  \Email{jzl0375@auburn.edu}\\
  \addr Department of Mathematics and Statistics\\
  Auburn University
  \AND
  \Name{Huan He}
  \Email{huan.he@auburn.edu}\\
  \addr Department of Mathematics and Statistics\\
  Auburn University
}

\begin{document}
\maketitle

\begin{abstract}
Augmenting small concurrent studies with external or historical cohorts
is attractive in drug development, where enrollment is slow, follow-up is
expensive, and closely related trial or real-world data are often
already available.
Bayesian dynamic borrowing (BDB) provides a principled framework for adaptively controlling the influence of external data, but classical implementations often depend on hand-specified priors and MCMC-based inference, which can be computationally expensive and not generalizable.
In this work, we study amortized neural posterior estimation (NPE) as a flexible alternative.
A single network is pretrained on simulated current/external dataset
pairs spanning covariate shift, outcome drift, and joint
non-exchangeability, and then returns an approximate posterior for a
scalar current-study target in a single forward pass.
Through simulation studies, we find that NPE is most useful under outcome drift and joint mismatch: in the harder outcome-drift regimes, it gives up to about five-fold lower absolute bias than the best classical baseline and keeps Type I error close to nominal. After pretraining, posterior summaries are obtained in about 8 ms per dataset, roughly $10^3\times$ faster than MCMC-based borrowing baselines in our timing experiment. We further analyze Alzheimer's Disease Neuroimaging Initiative (ADNI) data and show that, when mild cognitive impairment outcomes differ across cohorts, the NPE formulation recovers the later-cohort risk level in this example without claiming greater precision. Code is available at \url{https://github.com/ChinHungScott/NPE-for-Bayesian-Dynamic-Borrowing-MLHC-}.
\end{abstract}

\input{sections/001intro}
\input{sections/002related}
\input{sections/003problem_setup}
\input{sections/004simulation}
\input{sections/005adni}
\input{sections/006discussion}
\acks{Data used in preparation of this article were obtained from the
Alzheimer's Disease Neuroimaging Initiative (ADNI) database
(\url{adni.loni.usc.edu}). The ADNI was launched in 2003 as a
public--private partnership led by Principal Investigator Michael W.
Weiner, MD. ADNI data collection and sharing are funded by the National
Institute on Aging (National Institutes of Health Grant U19 AG024904) and
by additional public and private partners; the grantee organization is the
Northern California Institute for Research and Education, and private-sector
contributions are facilitated by the Foundation for the National Institutes
of Health. The ADNI investigators contributed to the design and implementation
of ADNI and/or provided data but did not participate in the analysis or
writing of this report. A complete listing of ADNI investigators is available at
\url{http://adni.loni.usc.edu/wp-content/uploads/how_to_apply/ADNI_Acknowledgement_List.pdf}.}
\newpage
\bibliography{refs}
\newpage
\appendix
\input{sections/007appendix}

\end{document}

%% file: sections/001intro.tex
\section{Introduction}
Small or slow-accruing clinical studies are a persistent bottleneck in
healthcare research.
Rare diseases, pediatric indications, and long-horizon neurodegenerative
cohorts all face the same problem: the concurrent study is
statistically underpowered on its own, yet substantial data from related
historical or external sources already exist.
Integrating external evidence with a concurrent study---through historical
control borrowing, synthetic control arms, or externally controlled
trials---has therefore become a central methodological and regulatory
question, with recent guidance from the U.S.\ FDA explicitly addressing the
design and conduct of externally controlled trials
\citep{fda2023external,thorlund2020synthetic,edwards2024using}.
In Alzheimer's disease (AD) research in particular, longitudinal follow-up is
expensive, clinically meaningful outcomes accrue slowly, and successive
cohorts often differ in enrollment criteria, disease severity, and assessment
practice \citep{petersen2010alzheimer,weiner2017alzheimer}.
Methods that borrow safely from related cohorts can make such studies more
informative, provided they avoid treating historical data as fully
exchangeable by default.

The key statistical object underlying these borrowing methods is
\emph{exchangeability}: the assumption that subject-level outcomes from the
external and concurrent sources are drawn from a common distribution, up to
covariates and random variation \citep{pocock1976combination}.
When exchangeability holds, pooling reduces variance and shrinks effective
sample-size requirements.
When it fails---because of secular trends, different diagnostic criteria,
population drift, or changes in the data-generating outcome model---naive
pooling induces bias and can inflate the type~I error rate of downstream
analyses \citep{viele2014use,koppschneider2020power}.
Bayesian dynamic borrowing (BDB) was developed to navigate this tension.
Methods such as the power prior \citep{ibrahim2000power,ibrahim2015power},
normalized and adaptive power priors
\citep{duan2006evaluating,ye2022normalized,gravestock2017adaptive}, the
commensurate prior \citep{hobbs2011commensurate,hobbs2012commensurate}, and
the (robust) meta-analytic-predictive prior
\citep{neuenschwander2010summarizing,schmidli2014robust} all attempt to
down-weight external data adaptively when prior--data conflict is detected
\citep{evans2006checking,presanis2013conflict}.

Despite their principled foundations and interpretation, existing BDB methods face several limitations for clinical trial data borrowing from real-world data (RWD). 
First, they summarize cross-source compatibility through one or a few
low-dimensional quantities---a scalar discount $a_0$, a commensurability
precision $\tau$, or a mixture weight $\epsilon$---and therefore cannot
cleanly separate covariate shift from outcome drift, even though these two
forms of non-exchangeability are handled very differently in practice
\citep{galwey2017supplementation,vanrosmalen2018including}.
Second, the discount hyperparameters themselves typically require tuning or
careful prior specification, and the resulting borrowing strength is
sensitive to those choices \citep{gravestock2017adaptive,viele2014use}.
Third, every new dataset pair requires a fresh posterior fit, usually via
MCMC, which is costly when many external-current comparisons must be run,
when operating characteristics must be simulated, or when borrowing decisions
must be reassessed for multiple candidate cohorts.
Reweighting alternatives---propensity-score matching, inverse-probability
weighting, overlap weighting, and entropy balancing
\citep{rosenbaum1983central,stuart2010matching,li2018balancing,
li2019addressing,hainmueller2012entropy}---address covariate shift well but
do not correct shifts in the outcome mechanism itself
\citep{dahabreh2019generalizing}, which is precisely the failure mode we
observe between ADNI1 and ADNIGO/ADNI2.

Amortized neural posterior estimation (NPE) offers a different route that
directly targets the computational and representational limits of classical BDB.
In NPE, a neural density estimator is pretrained on simulated
parameter--data pairs so that, at inference time, it returns an approximate
posterior in a single forward pass
\citep{papamakarios2016fast,papamakarios2019sequential,greenberg2019automatic,
cranmer2020frontier,lueckmann2021benchmarking,papamakarios2021normalizing}.
Once trained, the same network can be reused across many current/external
dataset pairs without re-running MCMC, and the representation fed into the
network can encode richer cross-source summaries than a single compatibility
scalar.
The design question specific to borrowing is therefore not whether NPE can
approximate a posterior, but how to represent the two sources so that the
network sees both the target estimand and the structure of
non-exchangeability, and how the resulting approximate posteriors behave
under realistic cohort shift.

In this paper we study a summary-based NPE approach for Bayesian dynamic
borrowing under distributional shift.
The method is pretrained on simulated current/external dataset pairs that
span a controlled grid of exchangeability regimes---including covariate
shift, outcome drift, and joint non-exchangeability---and is then reused
across dataset pairs with a single forward pass.
Our empirical claims concern a scalar current-study target: an additive
study-effect parameter in the Gaussian simulations and a cohort-level risk
parameter in the ADNI example. More complex targets, such as covariate
effects or treatment--covariate interactions, are important extensions but
are not part of the main validation; Appendix~\ref{app:covariate-effect}
reports a limited illustrative check for one covariate-effect target.
We first study the method in a controlled simulation setting spanning six
exchangeability regimes and compare against representative dynamic
borrowing and reweighting baselines.
We then present a real-data borrowing case study using the Alzheimer's
Disease Neuroimaging Initiative (ADNI)
\citep{petersen2010alzheimer,weiner2017alzheimer}: ADNI1 is treated as an
external cohort and ADNIGO/ADNI2 as a concurrent cohort among baseline mild
cognitive impairment (MCI) subjects, with 24-month conversion to dementia as
the outcome.
In this setting, the two cohorts have different event rates,
making naive borrowing poorly calibrated; the ADNI example is used as a
real-data borrowing problem under cohort shift rather than as a randomized
treatment-effect analysis.

\begin{figure}[H]
    \centering
    \includegraphics[width=.80\textwidth]{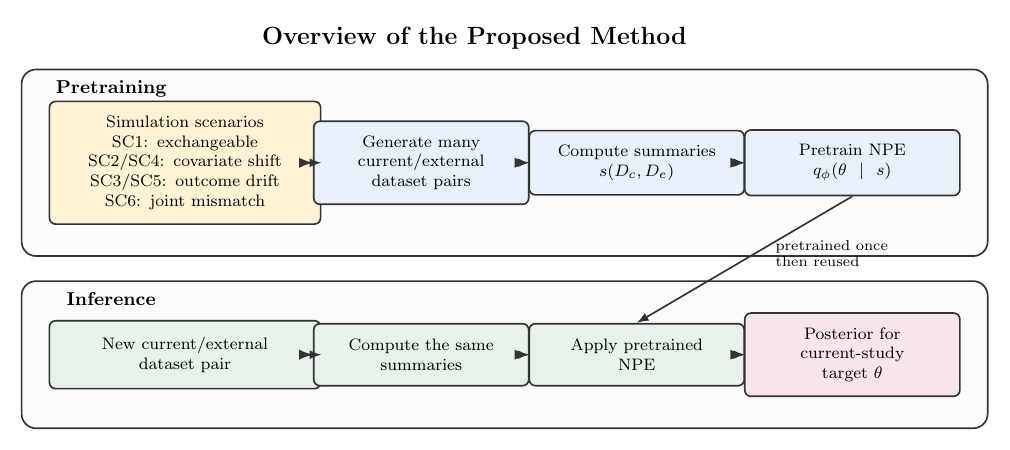}
    \caption{Proposed NPE workflow. Simulation scenarios generate
    current/external pairs for pretraining; at inference, the same summaries
    from a new pair are passed through the pretrained NPE to return a posterior
    for the current-study target.}
    \label{fig:method-overview}
\end{figure}

Our main contributions are as follows.
\begin{itemize}[leftmargin=1.5em, itemsep=2pt]
  \item We develop a summary-based NPE method for Bayesian dynamic
        borrowing that replaces repeated per-dataset posterior fitting with
        amortized inference on paired current/external data, so that a
        single pretrained network can be reused across many cohort pairs.
  \item We evaluate the method in a controlled simulation study spanning
        six exchangeability regimes, including covariate shift, outcome
        drift, and joint non-exchangeability, and benchmark against
        representative BDB priors and propensity-based reweighting
        baselines.
  \item We show empirically that the main strength of the approach is
        robustness in the harder mismatch settings, especially when the
        outcome model differs across sources---the regime where low-dimensional
        compatibility summaries and covariate-balancing weights tend to
        struggle.
  \item We present a real-data ADNI case study showing that this borrowing
        problem also appears in practice and that the proposed NPE
        formulation can recover the concurrent-cohort risk level more
        closely than naive borrowing strategies in this example.
\end{itemize}

The remainder of the paper is organized as follows.
Section~\ref{sec:related} reviews dynamic borrowing, balancing methods, and
simulation-based inference.
Section~\ref{sec:problem} introduces the borrowing setup, proposed method,
and simulation regimes.
Section~\ref{sec:simulation} presents the simulation study.
Section~\ref{sec:adni} reports the ADNI real-data example.
Section~\ref{sec:discussion} concludes with limitations and future directions.

\subsection*{Generalizable Insights about Machine Learning in the Context
             of Healthcare}

\begin{itemize}[leftmargin=1.5em, itemsep=2pt]
  \item \textbf{Amortized inference can make adaptive borrowing practical in
        healthcare studies with repeated external comparisons.}
        Pretraining an NPE model allows approximate posterior inference
        for new cohort pairs without rerunning expensive Bayesian computation
        from scratch each time.

  \item \textbf{Real-world borrowing problems often appear first as calibration
        mismatch rather than discrimination failure.}
        In the ADNI application, external-only models rank subjects well but
        systematically overestimate risk in the concurrent cohort, showing that
        borrowing decisions should account for cohort shift even when predictive
        AUC remains high.

  \item \textbf{A combined simulation-plus-case-study design is useful for
        evaluating healthcare borrowing methods.}
        Controlled simulations make it possible to assess bias and error under
        known exchangeability violations, while a real-world case study checks
        whether the same issues appear in practice and whether the method
        behaves reasonably on clinically meaningful data.
\end{itemize}

%% file: sections/002related.tex
\section{Related Work}
\label{sec:related}

\subsection{Bayesian dynamic borrowing and the exchangeability assumption}

Integrating external information into a concurrent study has a long history
in clinical biostatistics, beginning with the hierarchical framework of
\citet{pocock1976combination} for combining randomized and historical
controls.
Let $D_c$ denote the data from a small current study and $D_e$ the data from
an external or historical source, with a current-study target parameter
$\theta$.
The central statistical assumption underlying any borrowing procedure is
\emph{exchangeability} between the two sources: that, conditional on
covariates, observations from $D_c$ and $D_e$ share a common generative
mechanism.
When exchangeability holds, pooling reduces variance and improves
precision, yielding effective sample-size gains that are especially valuable
in rare-disease, pediatric, and oncology trials where concurrent accrual is
limited \citep{berry2013bayesian,ventz2019design,edwards2024using}.
When it fails---because of secular drift, differing diagnostic criteria, or
changes in the outcome mechanism---naive pooling induces bias and can inflate
type~I error \citep{viele2014use,koppschneider2020power,psioda2019bayesian}.
This tension motivates \emph{dynamic} rather than fixed borrowing: methods
that quantify cross-source compatibility and modulate the contribution of
$D_e$ accordingly.
Existing dynamic borrowing methods mainly differ in how they assess
compatibility between $D_c$ and $D_e$ and how that assessment is translated
into borrowing strength.

\paragraph{Power prior.}
The power prior \citep{ibrahim2000power,ibrahim2015power} incorporates
external information by discounting the external likelihood through a scalar
weight $a_0 \in [0,1]$:
\begin{equation}
p(\theta \mid D_c, D_e, a_0)
\;\propto\;
L(\theta; D_c)\,L(\theta; D_e)^{a_0}\,\pi(\theta).
\end{equation}
When $a_0 = 1$ the two sources are fully pooled, whereas $a_0 = 0$ ignores
the external data entirely.
Fixing $a_0$ a priori is difficult in practice.
The normalized power prior \citep{duan2006evaluating,ye2022normalized} treats
$a_0$ as unknown and integrates over it, and empirical-Bayes adaptive
variants \citep{gravestock2017adaptive} plug in a data-driven estimate.

\paragraph{Commensurate prior.}
The commensurate prior \citep{hobbs2011commensurate,hobbs2012commensurate}
introduces separate parameters for the current and external sources and
links them through a precision parameter $\tau$:
\begin{align}
\theta_e &\sim \pi(\theta_e), &
\theta_c \mid \theta_e, \tau &\sim \mathcal{N}(\theta_e,\tau^{-1}).
\end{align}
Large $\tau$ encourages $\theta_c$ and $\theta_e$ to be close, whereas small
$\tau$ allows them to differ.
With an appropriate hyperprior on $\tau$, the posterior adaptively reduces
borrowing under prior--data conflict \citep{evans2006checking,
presanis2013conflict}.

\paragraph{Meta-analytic-predictive and robust MAP priors.}
The meta-analytic-predictive (MAP) prior
\citep{neuenschwander2010summarizing} derives an informative prior for
$\theta_c$ by treating historical trials as exchangeable draws from a common
hierarchical model.
Because exchangeability can fail in ways that are hard to anticipate,
\citet{schmidli2014robust} proposed a robust MAP prior that mixes the
informative component with a vague one:
\begin{equation}
\pi_{\text{robust}}(\theta_c)
=
(1-\epsilon)\,\pi_{\text{MAP}}(\theta_c)
+
\epsilon\,\pi_{\text{vague}}(\theta_c),
\end{equation}
where $\epsilon$ controls protection against conflict between historical and
current data.
This construction is attractive in practice because $\epsilon$ has a
transparent interpretation, but the choice of $\epsilon$ and of the vague
component materially affects the resulting borrowing behavior.

\paragraph{Limitations of classical BDB.}
Several studies have documented recurring practical limitations of these
methods.
First, operating characteristics are sensitive to the discount
hyperparameters $(a_0,\tau,\epsilon)$, and strict type~I error control
essentially eliminates the power gains that motivated borrowing in the first
place \citep{koppschneider2020power,galwey2017supplementation,
vanrosmalen2018including}.
Second, classical BDB methods summarize cross-source compatibility through
\emph{one or a few low-dimensional quantities}, which can mask clinically
important source differences: in particular, covariate shift and outcome
drift enter the same scalar $a_0$ or $\tau$, even though they have very
different implications for bias in the current-study estimand
\citep{viele2014use,dahabreh2019generalizing}.
Third, these models require repeated posterior fitting---typically via
MCMC---for each new dataset pair, which is expensive when operating
characteristics must be simulated across many scenarios, when borrowing
decisions must be reassessed for several candidate external cohorts, or when
the analysis must be rerun as data accrue.
These limitations motivate borrowing procedures that can (i) exploit richer,
multi-dimensional summaries of source compatibility, and (ii) amortize
inference cost across many current/external pairs.

\subsection{Balancing and reweighting methods}

A related class of methods approaches external-data use through balancing or
reweighting rather than hierarchical priors.
Propensity-score matching and weighting
\citep{rosenbaum1983central,stuart2010matching} adjust the external cohort so
that its covariate distribution better matches the current cohort.
Inverse-probability weighting is a standard implementation but can be
unstable when estimated propensity scores are close to 0 or 1
\citep{chesnaye2022introduction}, and overlap weighting
\citep{li2018balancing,li2019addressing} addresses this instability by
emphasizing subjects in the region of covariate overlap, with weights
proportional to $2\hat e(x)(1-\hat e(x))$.
Entropy balancing \citep{hainmueller2012entropy} reweights the external
cohort to match prespecified moments of the current cohort without estimating
a propensity model.
These ideas extend naturally to transportability and generalizability of
causal estimands across populations
\citep{dahabreh2019generalizing}.
Regulators have also begun to formalize the use of external and synthetic
controls in pivotal studies \citep{thorlund2020synthetic,fda2023external}.

A key limitation of weighting-based approaches, however, is that they
primarily address differences in baseline \emph{covariates} but not
differences in the \emph{outcome} generation mechanism
\citep{viele2014use,dahabreh2019generalizing}.
In Alzheimer's cohorts in particular, an external and a current source can
also differ in recruitment period, diagnostic practice, assessment schedule,
and baseline conversion risk.
In that case, balancing baseline covariates may not be enough: even a
well-calibrated external model may encode a shifted outcome relationship
relative to the current cohort.
This calibration-level mismatch, rather than covariate imbalance, is the
dominant failure mode in our ADNI case study.

\subsection{Neural posterior estimation and simulation-based inference}

Simulation-based inference (SBI) is a family of methods that perform
Bayesian inference using only the ability to simulate from the generative
model \citep{cranmer2020frontier}.
Neural posterior estimation (NPE) is the branch of SBI that trains a
conditional density estimator $q_{\phi}(\theta \mid x)$ directly on
simulated parameter--data pairs
\citep{papamakarios2016fast,papamakarios2019sequential,greenberg2019automatic,
lueckmann2021benchmarking}, with training objective
\begin{equation}
\mathcal{L}(\phi)
=
\mathbb{E}_{\theta \sim \pi(\theta),\, x \sim p(x\mid \theta)}
\left[-\log q_{\phi}(\theta \mid x)\right].
\end{equation}
Once trained, the network can produce an approximate posterior for a new
observation in a single forward pass, avoiding repeated posterior sampling.
Normalizing flows are the standard choice for $q_\phi$ because they provide
flexible, exactly-evaluable densities
\citep{papamakarios2021normalizing}, and open-source toolkits such as
\texttt{sbi} \citep{tejerocantero2020sbi} and \texttt{BayesFlow}
\citep{radev2022bayesflow} have made these methods practical for scientific
applications ranging from neuroscience \citep{goncalves2020training} to
epidemiology.

Two features of NPE are directly relevant to dynamic borrowing.
First, inference is \emph{amortized}: the upfront simulation-and-training
cost can be reused across many new dataset pairs, which matches the setting
where the same borrowing procedure must be evaluated repeatedly as cohorts
change.
Second, the network's input can be any summary of the two sources, not just
a single compatibility scalar---so long as the simulator covers the
relevant exchangeability regimes.
This expands the representational budget that governs how borrowing strength
reacts to different kinds of cohort shift.

For dynamic borrowing, the representation supplied to the posterior
estimator must therefore encode both information about the target estimand
and information about \emph{source compatibility}.
This requirement differs from many standard SBI settings, in which the
primary challenge is compressing data without losing information about a
single target parameter.
In borrowing problems, the representation must additionally preserve
structure that is informative about when external data should and should
not influence inference---the perspective we take in the remainder of the
paper.

%% file: sections/003problem_setup.tex
\section{Method}
\label{sec:problem}

\subsection{Setup and inferential target}

Let $D_c = \{(x_{ci}, y_{ci})\}_{i=1}^{n_c}$ and
$D_e = \{(x_{ej}, y_{ej})\}_{j=1}^{n_e}$ denote the concurrent and
external datasets. Inference targets a scalar current-study parameter
$\theta$ defined on the \emph{concurrent} population; the external sample
enters the analysis only to the extent that it is compatible with that
population. In the Gaussian simulations, $\theta$ is an additive
current-study effect; in the ADNI example, $\theta$ is a cohort-level
logit-risk parameter.
Classical Bayesian dynamic borrowing formalizes this qualification through
a borrowing-strength parameter $\phi$---the power-prior discount $a_0$,
the commensurability precision $\tau$, or an analogous quantity---and draws
from the joint posterior $p(\theta, \phi \mid D_c, D_e)$.

The joint posterior $p(\theta, \phi \mid D_c, D_e)$ is not a single
object but a family indexed by modeling choices: the hyperprior on
$\phi$ (e.g., a $\mathrm{Beta}(a,b)$ on $a_0$, a half-normal scale on
$\tau$, or the mixture weight $\epsilon$ in a robust MAP prior), the
prior on $\theta$, and the parameterization of the borrowing component.
Responsible reporting (required by FDA) therefore typically requires sweeping over
several of these
\citep{schmidli2014robust,gravestock2017adaptive,vanrosmalen2018including},
and MCMC must be rerun for each combination as well as for each new
cohort pair or prior parameter.
Our runtime study in Section~\ref{sec:simulation} shows that a single
joint $(\theta, \phi)$ fit takes seconds under good mixing and longer
otherwise, and this cost multiplies across sensitivity settings,
cohorts, and subgroups.

\subsection{Amortized posterior inference}

Rather than re-solving the posterior per configuration, we train a single
conditional density estimator that maps any plausible
$(D_c, D_e)$ pair to an approximate posterior for $\theta$:
\begin{equation}
  q_{\eta}\!\bigl(\theta \mid s(D_c, D_e)\bigr)
  \;\approx\;
  p\!\bigl(\theta \mid D_c, D_e\bigr),
  \label{eq:amortized}
\end{equation}
where $s(\cdot,\cdot) \in \mathbb{R}^d$ is a fixed-length summary of the
two datasets and $\eta$ are the parameters of a neural network.
Training proceeds by simulation: given a prior $\pi(\theta, \psi)$ over the
target and nuisance parameters, a simulator
$(\theta, \psi) \mapsto (D_c, D_e)$, and $M$ simulated triples
$(\theta^{(m)}, D_c^{(m)}, D_e^{(m)})$, the parameters $\phi$ are
fit once. At deployment, inference for a new pair
$(D_c, D_e)$ reduces to computing $s$ and a single forward pass;
the per-dataset inference cost is independent of $M$ and does not require additional
sampling.
The approximation \eqref{eq:amortized} is valid only over the support of
the training simulator, so its fidelity depends directly on how broadly
the simulator covers plausible mismatch patterns (Section~\ref{sec:regimes}).

Amortization also aligns well with the structure of the borrowing problem.
The degree of borrowing is not a free analyst-specified quantity: it is
determined by the compatibility between $D_c$ and $D_e$. Classical
methods infer this compatibility implicitly through the posterior on $\phi$.
Here it is represented indirectly through $s$: if the simulator spans
regimes from full exchangeability to strong mismatch, the network can
learn how much to borrow as a function of observed mismatch.
The resulting $q_{\eta}(\theta \mid s)$ approximates the Bayes rule
\emph{induced by the training simulator}, not a universally optimal one;
the simulator's coverage is therefore part of the method rather than an
afterthought.

We use a Gaussian posterior family,
$
  q_{\eta}(\theta \mid s) = \mathcal{N}\!\bigl(m_{\eta}(s),\, \sigma_{\eta}^2(s)\bigr),
$
trained by minimizing the expected negative log-likelihood
\begin{equation}
  \widehat{\eta}
  \;=\;
  \arg\min_{\eta}\,
  \frac{1}{M}\sum_{m=1}^{M}
    -\log q_{\eta}\!\bigl(\theta^{(m)} \,\big|\, s(D_c^{(m)}, D_e^{(m)})\bigr).
  \label{eq:training-obj}
\end{equation}
Two remarks on this choice.
First, \eqref{eq:training-obj} is the standard NPE objective; as
$M \to \infty$ and for sufficiently expressive $q_\eta$, its minimizer
converges to $p(\theta \mid s)$
\citep{papamakarios2016fast,greenberg2019automatic}.
Second, the Gaussian head is a deliberate simplification.
Posteriors for a scalar target under weakly informative priors
and moderate sample sizes are typically close to Gaussian, and in
preliminary experiments richer density families standard in NPE
(normalizing flows, mixture heads) did not yield measurable gains in this
setting while introducing additional calibration concerns. We therefore
default to the Gaussian head.
Architecture and training details are reported in
Appendix~\ref{app:implementation}.

\subsection{Summary statistics}
\label{sec:summary-stats}

Because the network sees the data only through $s(D_c, D_e)$, the design
of $s$ is the main modeling lever.
We aim for a summary that is approximately sufficient for the mismatch
structure relevant to $\theta$: pairs $(D_c, D_e)$ that should yield the
same posterior on $\theta$ should, under the simulator, yield similar $s$.
In addition to the original data space, we use a summary assembled from four blocks, each
targeting a distinct facet of source (in)compatibility:
\begin{enumerate}
  \item \textbf{Marginal moments.} Per-source means and standard
    deviations of $X$ and $Y$, together with the sample sizes
    $(n_c, n_e)$. Intended to detect first- and second-moment shifts.
  \item \textbf{Conditional structure.} OLS coefficients
    $\hat{\beta}_c, \hat{\beta}_e$ and residual variances obtained by
    fitting $Y \sim X$ separately in each source.
    The gap $\Delta_\beta = \hat\beta_c - \hat\beta_e$ is intended as a
    signal of outcome-mechanism drift that marginal summaries cannot see.
  \item \textbf{Covariate overlap.} Propensity-score summaries based on
    $e(x) = P(S{=}1 \mid X{=}x)$, where $S{=}1$ indexes the concurrent
    source. External observations carry overlap weights
    $w_j = 2\sqrt{e(x_{ej})\{1 - e(x_{ej})\}}$, from which we compute a
    weighted external mean, the effective sample size
    $n_e^{\mathrm{eff}} = (\sum_j w_j)^2 / \sum_j w_j^2$, and a weight
    concentration score. These summarize propensity-score overlap, the
    quantity used by propensity-based borrowing baselines.
  \item \textbf{Joint-shift diagnostic.} The per-covariate mean gap
    $\Delta_\mu = \bar{x}_c - \bar{x}_e$ and the Wasserstein-1 distance
    between the empirical covariate distributions. Intended to capture
    joint mismatch that marginal or conditional summaries alone may not
    resolve.
\end{enumerate}
The blocks are designed to be partially redundant in exchangeable regimes
and complementary under mismatch: under exchangeability all four blocks
are expected to read small values, covariate shift is reflected primarily
in Blocks~3--4, outcome drift primarily in Block~2, and joint mismatch in
Blocks~2 and 4 together. We refer to this block structure as
\emph{exchangeability-aware} in the sense that the summary is organized by
type of mismatch rather than by a single compatibility scalar;
whether the network makes effective use of this structure is an empirical
question that we examine in Section~\ref{sec:simulation}.
This organization also gives a simple auditing layer: before trusting a
posterior, an analyst can inspect whether the observed pair is unusual in
the marginal-moment block, the outcome-mechanism block, the overlap block,
or the joint-shift block. The neural network itself is not intrinsically
interpretable, and if the relevant mismatch is absent from these summaries,
posterior quality can degrade; the intent is to make the main mismatch
signals explicit rather than hidden in an opaque latent representation.
Appendix~\ref{app:ols-perturbation} reports a perturbation check for the
OLS-derived coefficient-difference block.

\subsection{Pretraining via simulated borrowing scenarios}
\label{sec:regimes}

Because \eqref{eq:amortized} holds only on the support of the training
distribution, pretraining is part of the method rather than a
preprocessing step: $q_\eta$ is only as informed about mismatch as the
simulator exposes.
The simulator must produce paired datasets $(D_c, D_e)$ across a
spectrum of mismatch conditions together with the $\theta$ that
generated them; drawing $M$ such triples and minimizing
\eqref{eq:training-obj} yields a network reusable across downstream
analyses.

This simulation step is not fit to the analysis data. It plays the role of a
prior-generating model: the analyst specifies plausible ranges for the target,
nuisance parameters, covariate shift, and outcome drift, then trains the NPE
on data pairs drawn from that range. At inference time the pretrained network
sees only the observed summaries $s(D_c,D_e)$ from the real current/external
pair. Thus, as in any Bayesian borrowing analysis, the key modeling
assumption is that the observed pair lies within the support of the prior
predictive distribution induced by the simulator. These ranges can be elicited
from domain knowledge, published cohort summaries, or deliberately conservative
bounds when little prior evidence is available; they are not tuned to the
observed target answer.

We specify the simulator at the level of \emph{what mismatch must be
represented}, not a particular data-generating form.
Two axes---covariate exchangeability and outcome exchangeability---cover
the borrowing problems we consider, yielding the six regimes of
Table~\ref{tab:regimes}. During pretraining we sample these two axes
\emph{continuously} from broad priors spanning SC1 through SC6, so
$q_\eta$ learns a smooth map from observed mismatch to posterior; the
discrete grid is reserved for interpretable evaluation
(Section~\ref{sec:simulation}).

\begin{table}[t]
\centering
\small
\begin{tabular}{cll}
\toprule
Scenario & Covariates & Outcomes \\
\midrule
SC1 & exchangeable        & exchangeable \\
SC2 & partial shift       & exchangeable \\
SC3 & exchangeable        & partial drift \\
SC4 & strong shift        & exchangeable \\
SC5 & exchangeable        & strong drift \\
SC6 & strong shift        & strong drift \\
\bottomrule
\end{tabular}
\caption{Two-axis taxonomy of source-mismatch regimes. SC1 is the
exchangeable baseline; SC2 and SC4 isolate covariate shift at increasing
strength; SC3 and SC5 isolate outcome drift at increasing strength; and SC6
combines strong covariate shift with strong outcome drift.}
\label{tab:regimes}
\end{table}

The concrete data-generating form is application-dependent: a Gaussian
linear model for the simulation study (Section~\ref{sec:simulation})
and a binary-outcome model calibrated to the cohort for ADNI
(Section~\ref{sec:adni}). The same $q_\eta$ pipeline applies in both
cases; only the simulator, and hence the pretrained weights, differ.

%% file: sections/004simulation.tex
\section{Simulation Study}
\label{sec:simulation}

\subsection{Design}
We evaluate the method on the two-axis taxonomy of mismatch regimes
introduced in Section~\ref{sec:regimes} (Table~\ref{tab:regimes}),
instantiated as a Gaussian linear data-generating model:
\begin{align}
  X_{ci} &\sim \mathcal{N}(\mu_c \mathbf{1}_p,\, \Sigma), &
  X_{ej} &\sim \mathcal{N}(\mu_e \mathbf{1}_p,\, \Sigma), \\
  Y_{ci} &= X_{ci}^\top \beta + \theta + \varepsilon_{ci}, &
  Y_{ej} &= X_{ej}^\top \beta + \theta + \delta_e + \varepsilon_{ej},
\end{align}
with $\varepsilon \sim \mathcal{N}(0, \sigma^2)$.
Covariate shift is controlled by $\mu_e - \mu_c$ and outcome drift by
$\delta_e$, matching the two axes of Table~\ref{tab:regimes}:
SC1 sets both to zero, SC2 and SC4 set $\mu_e-\mu_c$ to partial and
strong levels with $\delta_e=0$, SC3 and SC5 set $\delta_e$ to partial
and strong levels with $\mu_e=\mu_c$, and SC6 sets both to strong
levels. Training draws $(\mu_e-\mu_c, \delta_e)$ continuously over the
full range that these anchor points span, so SC1--SC6 are evaluation
points rather than training conditions. We use this grid as a scenario-level
ablation: each regime turns on a specific mismatch axis, or their
combination, so changes in bias, coverage, and Type I error can be attributed
to the kind of non-exchangeability present.
For each regime, we generate one concurrent and one external dataset.
We consider $N_c \in \{50,100\}$ and fix $N_e=200$.
The target parameter is the additive current-study target $\theta$.

\subsection{Methods compared}
We compare four methods:
\textbf{PSPower} \citep{ibrahim2015power}: a
    propensity-score-stratified power prior with a borrowing weight estimated
    within PS strata.
  \textbf{IW} \citep{chesnaye2022introduction}: an individualized
    weighting method based on covariate distance from the concurrent
    distribution.
\textbf{Commensurate} \citep{hobbs2011commensurate}: a
    commensurate prior whose precision parameter adapts to observed
    source compatibility.
\textbf{NPE} (ours): neural posterior estimation trained
    once on simulated datasets and reused to approximate
    $p(\theta \mid \mathcal{D}_c, \mathcal{D}_e)$ for new dataset pairs.

We expect PSPower and Commensurate to work best near exchangeability, IW to
help mainly under covariate shift, and NPE to be most useful when mismatch
involves outcome drift or joint covariate/outcome shift.

\subsection{Evaluation metrics}

We use two simulation settings.
In the non-null setting, we set $\theta=1$ and report bias, RMSE, coverage,
and power.
Bias and RMSE measure point estimation error.
Coverage checks whether the 95\% interval contains the true value.
Power is the fraction of 95\% intervals that exclude 0.

In the null setting, we set $\theta=0$ and report coverage and Type I error.
Type I error is the fraction of 95\% intervals that exclude 0 when the true
effect is zero.
The main text focuses on bias, coverage, and Type I error because these most
directly show the borrowing failure modes under source mismatch; RMSE and power
are included in the full simulation tables in Appendix~\ref{app:additional-results}.

\subsection{Main results for the non-null setting}

The non-null run with $\theta=1$ shows the clearest separation under outcome
drift. In SC1 and SC2, the classical methods already perform well and NPE is
somewhat more variable. In SC3 and SC5, however, the classical borrowing
methods have large bias and essentially no coverage, while NPE remains closer
to the current-study target. In SC6, NPE is more stable than PSPower and
Commensurate and is close to IW.

Table~\ref{tab:sim-hard-summary} summarizes the harder settings for
$N_c=100$; full simulation tables are in Appendix~\ref{app:additional-results}.
The MCSE columns show that the large method differences are not simulation
noise.

\vspace{-0.25em}
\begin{center}
\fontsize{7.25pt}{7.95pt}\selectfont
\renewcommand{\arraystretch}{0.88}
\captionof{table}{Representative hard-setting summary ($N_c=100$). Bias and
coverage are from $\theta=1$; Type I error is from $\theta=0$.}
\label{tab:sim-hard-summary}
\vspace{0.05em}
\setlength{\tabcolsep}{1.65pt}
\begin{tabular}{@{}llcccccc@{}}
\toprule
Scenario & Method & Bias & Bias SE & Cov. & Cov. SE & Type I & Type I SE \\
\midrule
SC3 & PSPower      & 0.96 & 0.02 & 0.00 & 0.00 & 1.00 & 0.00 \\
SC3 & IW           & 0.66 & 0.02 & 0.00 & 0.00 & 1.00 & 0.00 \\
SC3 & Commensurate & 0.86 & 0.02 & 0.00 & 0.00 & 1.00 & 0.00 \\
SC3 & NPE          & 0.15 & 0.02 & 1.00 & 0.00 & 0.02 & 0.02 \\
\cmidrule(lr){1-8}
SC5 & PSPower      & 0.95 & 0.01 & 0.00 & 0.00 & 1.00 & 0.00 \\
SC5 & IW           & 0.66 & 0.02 & 0.00 & 0.00 & 1.00 & 0.00 \\
SC5 & Commensurate & 0.86 & 0.02 & 0.00 & 0.00 & 1.00 & 0.00 \\
SC5 & NPE          & 0.13 & 0.02 & 1.00 & 0.00 & 0.04 & 0.03 \\
\cmidrule(lr){1-8}
SC6 & PSPower      & 0.32 & 0.02 & 0.32 & 0.07 & 0.68 & 0.07 \\
SC6 & IW           & 0.07 & 0.02 & 0.92 & 0.04 & 0.08 & 0.04 \\
SC6 & Commensurate & 0.21 & 0.02 & 0.54 & 0.07 & 0.46 & 0.07 \\
SC6 & NPE          & 0.07 & 0.02 & 0.98 & 0.02 & 0.06 & 0.03 \\
\bottomrule
\end{tabular}
\renewcommand{\arraystretch}{1}
\end{center}

\begin{figure}[H]
\centering
\includegraphics[width=.50\textwidth]{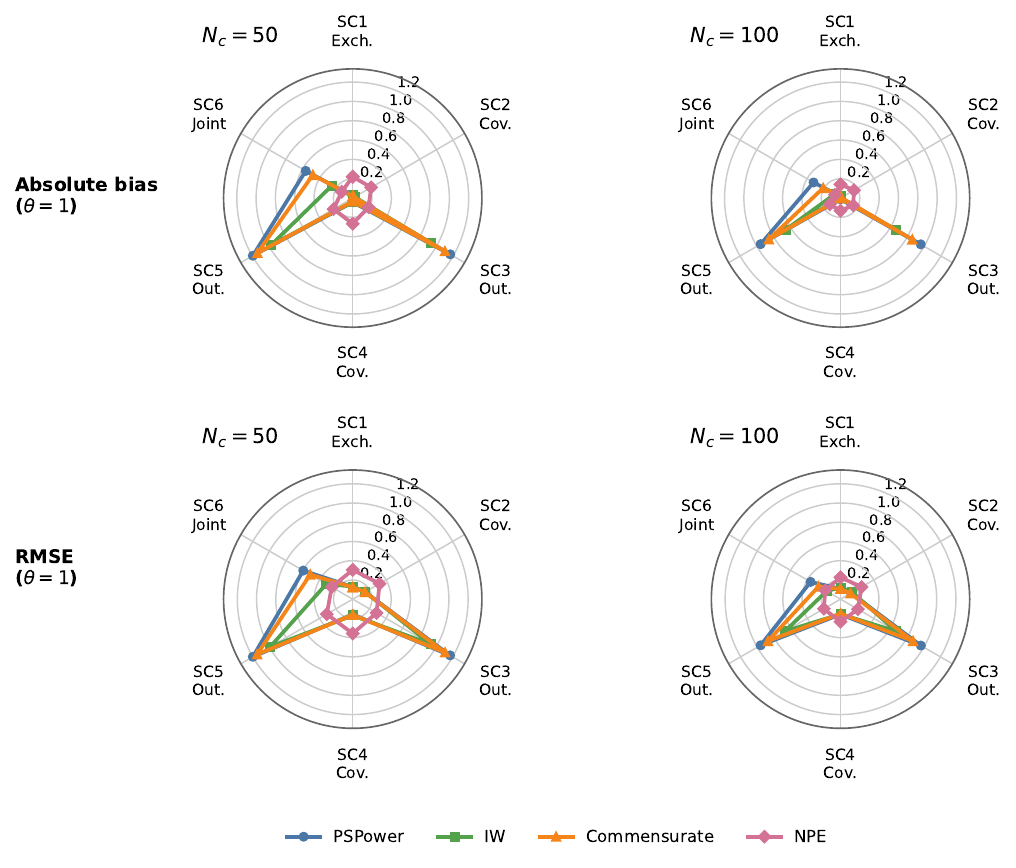}
\vspace{-0.4em}
\caption{Non-null estimation error by scenario. Each spoke is one
simulation scenario; values closer to the center indicate smaller error.}
\label{fig:sim-nonnull-error}
\end{figure}

Figure~\ref{fig:sim-nonnull-error} shows the same pattern across all
scenarios. Absolute bias measures displacement from the target, while RMSE
also reflects variability across simulation replicates.

\subsection{Null results}

The null run with $\theta=0$ gives a similar robustness pattern, now focused on
Type I error. In SC1 and SC2, the classical methods have cleaner operating
characteristics. In SC3, SC5, and SC6, they become anti-conservative, whereas
NPE stays much closer to nominal behavior.
The full null table is reported in Appendix~\ref{app:additional-results}.
Appendix Figure~\ref{fig:app-sim-null-operating} shows coverage and Type I
error in the null run.

\FloatBarrier
\subsection{Computational timing}
We benchmark NPE against two Bayesian dynamic-borrowing baselines that
admit a \emph{joint} posterior over the treatment effect and the borrowing-strength parameter: the Normalized Power Prior (NPP),
targeting $p(\theta, a_0 \mid \mathcal{D}_c, \mathcal{D}_e)$, and the
Commensurate Prior, targeting
$p(\theta, \tau \mid \mathcal{D}_c, \mathcal{D}_e)$. We restrict the
comparison to MCMC-based methods because only they target the same
inferential object---a joint distribution over $(\theta, \phi)$;
propensity-score and individualized-weights methods expose no $\phi$
and thus admit no ESS-matched criterion. For each method $\times$
scenario $\times$ replication ($B=5$ reps, four shift regimes
SC1/SC3/SC5/SC6, $n_c=100$, $n_e=200$), we run a doubling MCMC schedule
(initial length $500$, cap $32{,}000$ post-burnin) and record the
wall-clock at which
$\min\!\big(\mathrm{ESS}(\theta),\,\mathrm{ESS}(\phi)\big) \ge 1000$.
Posterior shape is diagnosed via the
Wasserstein-1 distance of the marginal $\theta$-posterior to a
long-chain NPP reference ($20{,}000$ post-burnin samples). NPE is
timed on its single amortized forward pass.

\begin{figure}[!htbp]
\centering
\includegraphics[width=.66\textwidth]{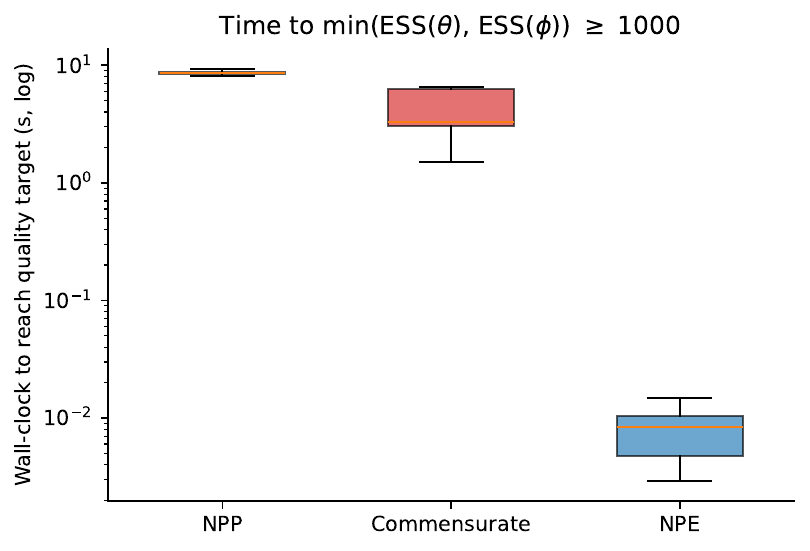}
\vspace{0.5em}
\caption{\textbf{Time to a high-quality joint posterior.}
Per-dataset wall-clock (log scale) required to reach
$\min(\mathrm{ESS}(\theta),\mathrm{ESS}(\phi)) \ge 1000$ for NPP and
Commensurate; NPE is timed on its single amortized forward pass.
Boxes span the IQR across $B=5$ replications $\times$ four
distribution-shift scenarios. NPE is $\sim\!10^3\times$ faster than
either MCMC sampler and uniformly so across regimes.}
\label{fig:time-to-quality}
\end{figure}

NPE returns its marginal-$\theta$ posterior in about 8 ms per dataset across
all scenarios. The Commensurate sampler requires 1.6--3.3 s (median) to meet
the ESS target, while the NPP sampler exhausts its 8.6 s budget without
reaching $\mathrm{ESS}\ge 1000$ in any replication. Its marginal
$\theta$-posterior is close to the long-chain reference, but the chain mixes
slowly because $\theta$ and $a_0$ are strongly correlated. Thus, NPP is
accurate but expensive to certify, whereas NPE gives a usable posterior summary
two-to-three orders of magnitude faster across shift regimes.

\subsection{Interpretation}
Overall, NPE is not uniformly best: in the easier regimes, classical borrowing
methods are often more efficient. Its main advantage appears when source
mismatch includes outcome drift or joint covariate/outcome shift, where fixed
borrowing rules can produce large bias, poor coverage, and inflated Type I
error.
\FloatBarrier

%% file: sections/005adni.tex
\section{Real-Data Example: ADNI}
\label{sec:adni}

We complement the simulation study with a real-data example based on the
Alzheimer's Disease Neuroimaging Initiative (ADNI). The purpose of this section
is different from that of the simulations. The simulations are used for
controlled methodological evaluation, whereas the ADNI analysis is meant to
show that the borrowing problem also appears clearly in an observed dataset.
Data were obtained from the ADNI database (\url{adni.loni.usc.edu}), a
public--private partnership launched in 2003 to study progression of mild
cognitive impairment and Alzheimer's disease. ADNI collects longitudinal
clinical, imaging, biomarker, and neuropsychological assessments to support
research on disease progression.

ADNI is not a randomized treatment study, so we do not treat this section as a
treatment-effect analysis. Instead, we use it as a cohort-difference risk
problem. The question is whether borrowing from an earlier ADNI cohort would
pull inference away from the later concurrent cohort, and whether the proposed
NPE formulation can better recover the concurrent risk level.

\subsection{ADNI setup}

We define the external cohort as ADNI1 and the concurrent cohort as the pooled
ADNIGO and ADNI2 samples. The analysis is restricted to subjects with mild
cognitive impairment (MCI) at baseline. The outcome is conversion to dementia
(or Alzheimer's disease) by approximately 24 months. For subject $i$, let
\[
Y_i =
\begin{cases}
1, & \text{if subject } i \text{ converts by approximately 24 months},\\
0, & \text{otherwise.}
\end{cases}
\]

For each subject, we use the baseline covariate vector
\[
X_i = (\text{age},\ \text{female},\ \text{education},\ \text{APOE4 count},\
\text{MMSE}_{\mathrm{bl}},\ \text{ADAS13}_{\mathrm{bl}},\
\text{CDRSB}_{\mathrm{bl}},\ \text{FAQ}_{\mathrm{bl}}).
\]

Table~\ref{tab:adni-cohort-summary} summarizes the two cohorts. The external
cohort has a much higher conversion rate than the concurrent cohort, making
naive borrowing questionable.

\begin{table}[H]
\centering
\caption{ADNI cohort summary for the real-data borrowing example.}
\label{tab:adni-cohort-summary}
\begin{tabular}{lcc}
\toprule
Cohort & Number of subjects & Conversion rate \\
\midrule
Concurrent (ADNIGO + ADNI2) & 376 & 0.1702 \\
External (ADNI1)            & 337 & 0.3769 \\
\bottomrule
\end{tabular}
\end{table}

\subsection{Baseline benchmark}

Before applying NPE, we first check whether the data show the borrowing issue
we are trying to study. We fit repeated logistic-regression baselines and
evaluate them on a held-out concurrent test set:  \textbf{ConcurrentOnly}: trained only on concurrent data;
     \textbf{ExternalOnly}: trained only on external data;
     \textbf{Pooled}: trained on the combined external and concurrent data.

The results are shown in Table~\ref{tab:adni-baselines}. All three approaches
have similar AUC and accuracy, but calibration differs. The
\texttt{ExternalOnly} and \texttt{Pooled} models predict risks that are too high
relative to the concurrent cohort, suggesting that direct borrowing mainly
distorts the risk level rather than discrimination.

\begin{table}[H]
\centering
\caption{Baseline benchmark on a held-out concurrent test set.}
\label{tab:adni-baselines}
\begin{tabular}{lcccccc}
\toprule
Method & AUC & Accuracy & Brier & LogLoss & MeanPred & MeanObs \\
\midrule
ConcurrentOnly & 0.918 & 0.881 & 0.084 & 0.271 & 0.183 & 0.168 \\
ExternalOnly   & 0.923 & 0.887 & 0.088 & 0.303 & 0.248 & 0.168 \\
Pooled         & 0.923 & 0.883 & 0.083 & 0.276 & 0.218 & 0.168 \\
\bottomrule
\end{tabular}
\end{table}

\subsection{ADNI-specific NPE formulation}

Since ADNI does not provide the same randomized treatment setting as the
simulation study, we use a simpler formulation in which the target is the
concurrent risk level. Let $X_c$ and $X_e$ denote the observed covariates in
the concurrent and external cohorts. We simulate outcomes according to
\begin{equation}
    Y_c \sim \mathrm{Bernoulli}\!\left(\sigma(X_c\beta + \theta)\right), \quad
Y_e \sim \mathrm{Bernoulli}\!\left(\sigma(X_e\beta + \theta + \delta)\right),
\end{equation}
where $\sigma(z)=1/(1+e^{-z})$ is the logistic link.

Here $\theta$ is the concurrent risk-level target, $\delta$ represents the
external--concurrent outcome shift, and $\beta$ denotes covariate effects.
The shift parameter allows the external cohort to differ rather than forcing
full exchangeability; it is used during pretraining but is not reported as a
target estimand.

To generate training data for NPE, we first fit a logistic model to the
observed ADNI data and use the fitted coefficient vector as an anchor
$\beta_{\mathrm{anchor}}$. We then simulate outcomes on the real ADNI
covariates by sampling
\[
\theta \sim \mathrm{Uniform}(-3,0), \qquad
\delta \sim N(0,0.75^2),
\]
adding small perturbations around $\beta_{\mathrm{anchor}}$, and generating
$Y_c$ and $Y_e$ on the fixed covariate matrices $X_c$ and $X_e$. For each
simulated dataset, we compute summary statistics based on covariate means and
standard deviations, event-rate summaries, between-cohort differences, and
sample-size information. The NPE model is then trained to approximate the
posterior distribution of $\theta$.

\subsection{ADNI results}

The posterior mean of the concurrent risk parameter is $-1.5990$, with
posterior standard deviation $0.8585$ and a 95\% interval of
$(-3.3009,\ 0.1082)$. The observed concurrent event rate is 0.1702, which
corresponds to an empirical concurrent logit risk of
$\mathrm{logit}(0.1702) \approx -1.5841$.
The NPE posterior mean is therefore very close to the empirical concurrent
target.

\begin{table}[H]
\centering
\caption{Main ADNI real-data result for the NPE-based formulation.}
\label{tab:adni-main-results}
\begin{tabular}{lc}
\toprule
Quantity & Value \\
\midrule
Posterior mean of $\theta$     & $-1.5990$ \\
Posterior SD of $\theta$       & $0.8585$ \\
95\% interval for $\theta$     & $(-3.3009,\ 0.1082)$ \\
Observed concurrent event rate & $0.1702$ \\
Observed external event rate   & $0.3769$ \\
Empirical concurrent logit     & $-1.5841$ \\
\bottomrule
\end{tabular}
\end{table}

This result is consistent with recovering the later-cohort risk level in this
example, but not with claiming greater precision. Appendix~\ref{app:adni-sensitivity}
checks sensitivity to the prior scale for the shift parameter $\delta$.

\FloatBarrier

\FloatBarrier

%% file: sections/006discussion.tex
\section{Discussion}
\label{sec:discussion}

We studied neural posterior estimation as an amortized approach to Bayesian
dynamic borrowing: pretrain once on a broad simulation distribution, then obtain
posteriors for new current--external pairs in a single forward pass. The
simulation study isolates covariate shift, outcome drift, and their combination;
ADNI illustrates that cohort mismatch can pull inference away from the
concurrent target even when discrimination remains strong. The empirical scope
is a scalar current-study target---an additive study-effect parameter in the
Gaussian simulations and a cohort-level risk parameter in ADNI. Covariate
effects, treatment--covariate interactions, and other causal functionals require
additional evaluation; Appendix~\ref{app:covariate-effect} provides only a
limited illustrative check for one target-aware covariate-effect analysis. This
framing keeps the real-data example aligned with the estimand actually analyzed
rather than treating ADNI as a randomized treatment-effect study.

Posterior quality depends on the pretraining simulator, so mismatch patterns
outside its support cannot be handled reliably. The current hand-crafted
summaries expose auditable signals of covariate shift, outcome-mechanism drift,
overlap, and joint mismatch, but may miss richer patient-level structure.
Runnable code is included in the supplementary material; ADNI raw data require
separate data-use access. Future work will study richer paired-data
representations, such as patient-level encoders for irregular longitudinal
records, posterior-spread calibration in severe mismatch regimes, multi-source
borrowing, and broader evaluation across estimands and prospective
data-borrowing settings.

%% file: sections/007appendix.tex
\section{Additional Results}
\label{app:additional-results}

\subsection{Full Simulation Tables}

Tables~\ref{tab:sim-full-nonnull} and \ref{tab:sim-full-null} report the full
simulation results corresponding to the main-text figures.

\scriptsize
\setlength{\tabcolsep}{3pt}
\renewcommand{\arraystretch}{1.08}
\begin{longtable}{llccccc}
\caption{Full non-null simulation results ($\theta = 1$).}
\label{tab:sim-full-nonnull}\\
\toprule
$N_c$ & Scenario & Method & Bias & RMSE & Coverage & Power \\
\midrule
\endfirsthead
\toprule
$N_c$ & Scenario & Method & Bias & RMSE & Coverage & Power \\
\midrule
\endhead
50 & SC1 & PSPower      & 0.0183 & 0.1268 & 0.90 & 1.00 \\
& SC1 & IW           & 0.0365 & 0.1309 & 0.94 & 1.00 \\
& SC1 & Commensurate & 0.0183 & 0.1234 & 0.90 & 1.00 \\
& SC1 & NPE          & 0.2201 & 0.3049 & 0.94 & 1.00 \\
\cmidrule(lr){2-7}
& SC2 & PSPower      & -0.0126 & 0.1485 & 0.94 & 1.00 \\
& SC2 & IW           & -0.0223 & 0.1505 & 0.98 & 1.00 \\
& SC2 & Commensurate & -0.0160 & 0.1464 & 0.94 & 1.00 \\
& SC2 & NPE          & 0.2202 & 0.3219 & 0.92 & 1.00 \\
\cmidrule(lr){2-7}
& SC3 & PSPower      & 1.1655 & 1.1693 & 0.00 & 1.00 \\
& SC3 & IW           & 0.9306 & 0.9369 & 0.00 & 1.00 \\
& SC3 & Commensurate & 1.1016 & 1.1091 & 0.00 & 1.00 \\
& SC3 & NPE          & 0.1889 & 0.2859 & 0.98 & 1.00 \\
\cmidrule(lr){2-7}
& SC4 & PSPower      & 0.0432 & 0.1579 & 0.94 & 1.00 \\
& SC4 & IW           & 0.0394 & 0.1697 & 0.98 & 1.00 \\
& SC4 & Commensurate & 0.0347 & 0.1630 & 0.94 & 1.00 \\
& SC4 & NPE          & 0.2661 & 0.3570 & 0.92 & 1.00 \\
\cmidrule(lr){2-7}
& SC5 & PSPower      & 1.1936 & 1.1993 & 0.00 & 1.00 \\
& SC5 & IW           & 0.9810 & 0.9915 & 0.00 & 1.00 \\
& SC5 & Commensurate & 1.1399 & 1.1488 & 0.00 & 1.00 \\
& SC5 & NPE          & 0.2298 & 0.3115 & 0.94 & 1.00 \\
\cmidrule(lr){2-7}
& SC6 & PSPower      & 0.5617 & 0.5943 & 0.08 & 1.00 \\
& SC6 & IW           & 0.2521 & 0.3167 & 0.68 & 1.00 \\
& SC6 & Commensurate & 0.4765 & 0.5107 & 0.16 & 1.00 \\
& SC6 & NPE          & 0.1311 & 0.2523 & 0.98 & 1.00 \\
\midrule
100 & SC1 & PSPower      & 0.0178 & 0.1063 & 0.94 & 1.00 \\
& SC1 & IW           & 0.0238 & 0.1205 & 0.98 & 1.00 \\
& SC1 & Commensurate & 0.0183 & 0.1058 & 0.92 & 1.00 \\
& SC1 & NPE          & 0.1413 & 0.2270 & 0.98 & 1.00 \\
\cmidrule(lr){2-7}
& SC2 & PSPower      & 0.0032 & 0.1261 & 0.94 & 1.00 \\
& SC2 & IW           & 0.0044 & 0.1422 & 0.94 & 1.00 \\
& SC2 & Commensurate & 0.0081 & 0.1195 & 0.94 & 1.00 \\
& SC2 & NPE          & 0.1610 & 0.2532 & 0.98 & 1.00 \\
\cmidrule(lr){2-7}
& SC3 & PSPower      & 0.9587 & 0.9651 & 0.00 & 1.00 \\
& SC3 & IW           & 0.6597 & 0.6701 & 0.00 & 1.00 \\
& SC3 & Commensurate & 0.8586 & 0.8714 & 0.00 & 1.00 \\
& SC3 & NPE          & 0.1518 & 0.2089 & 1.00 & 1.00 \\
\cmidrule(lr){2-7}
& SC4 & PSPower      & 0.0125 & 0.1576 & 0.86 & 1.00 \\
& SC4 & IW           & 0.0013 & 0.1552 & 0.90 & 1.00 \\
& SC4 & Commensurate & 0.0047 & 0.1497 & 0.90 & 1.00 \\
& SC4 & NPE          & 0.1312 & 0.2352 & 1.00 & 1.00 \\
\cmidrule(lr){2-7}
& SC5 & PSPower      & 0.9547 & 0.9597 & 0.00 & 1.00 \\
& SC5 & IW           & 0.6557 & 0.6681 & 0.00 & 1.00 \\
& SC5 & Commensurate & 0.8564 & 0.8710 & 0.00 & 1.00 \\
& SC5 & NPE          & 0.1299 & 0.1998 & 1.00 & 1.00 \\
\cmidrule(lr){2-7}
& SC6 & PSPower      & 0.3222 & 0.3595 & 0.32 & 1.00 \\
& SC6 & IW           & 0.0733 & 0.1672 & 0.92 & 1.00 \\
& SC6 & Commensurate & 0.2075 & 0.2633 & 0.54 & 1.00 \\
& SC6 & NPE          & 0.0668 & 0.1835 & 0.98 & 1.00 \\
\bottomrule
\end{longtable}
\normalsize

\scriptsize
\setlength{\tabcolsep}{3pt}
\renewcommand{\arraystretch}{1.08}
\begin{longtable}{llcccc}
\caption{Full null simulation results ($\theta = 0$).}
\label{tab:sim-full-null}\\
\toprule
Scenario & Method & Bias & RMSE & Coverage & Type I \\
\midrule
\endfirsthead
\toprule
Scenario & Method & Bias & RMSE & Coverage & Type I \\
\midrule
\endhead
\multicolumn{6}{l}{$N_c=50$} \\
SC1 & PSPower      & 0.0187 & 0.1269 & 0.90 & 0.10 \\
SC1 & IW           & 0.0373 & 0.1311 & 0.94 & 0.06 \\
SC1 & Commensurate & 0.0187 & 0.1235 & 0.90 & 0.10 \\
SC1 & NPE          & 0.3172 & 0.3811 & 0.90 & 0.10 \\
\cmidrule(lr){1-6}
SC2 & PSPower      & -0.0117 & 0.1485 & 0.94 & 0.06 \\
SC2 & IW           & -0.0212 & 0.1503 & 0.98 & 0.02 \\
SC2 & Commensurate & -0.0152 & 0.1463 & 0.94 & 0.06 \\
SC2 & NPE          & 0.2954 & 0.3703 & 0.90 & 0.10 \\
\cmidrule(lr){1-6}
SC3 & PSPower      & 1.1659 & 1.1698 & 0.00 & 1.00 \\
SC3 & IW           & 0.9314 & 0.9377 & 0.00 & 1.00 \\
SC3 & Commensurate & 1.1022 & 1.1096 & 0.00 & 1.00 \\
SC3 & NPE          & 0.3333 & 0.4062 & 0.92 & 0.08 \\
\cmidrule(lr){1-6}
SC4 & PSPower      & 0.0441 & 0.1582 & 0.96 & 0.04 \\
SC4 & IW           & 0.0406 & 0.1700 & 0.98 & 0.02 \\
SC4 & Commensurate & 0.0355 & 0.1632 & 0.94 & 0.06 \\
SC4 & NPE          & 0.3495 & 0.4082 & 0.86 & 0.14 \\
\cmidrule(lr){1-6}
SC5 & PSPower      & 1.1941 & 1.1997 & 0.00 & 1.00 \\
SC5 & IW           & 0.9817 & 0.9923 & 0.00 & 1.00 \\
SC5 & Commensurate & 1.1404 & 1.1493 & 0.00 & 1.00 \\
SC5 & NPE          & 0.4052 & 0.4649 & 0.84 & 0.16 \\
\cmidrule(lr){1-6}
SC6 & PSPower      & 0.5626 & 0.5952 & 0.08 & 0.92 \\
SC6 & IW           & 0.2533 & 0.3176 & 0.68 & 0.32 \\
SC6 & Commensurate & 0.4774 & 0.5114 & 0.18 & 0.82 \\
SC6 & NPE          & 0.2874 & 0.3508 & 0.92 & 0.08 \\
\midrule
\multicolumn{6}{l}{$N_c=100$} \\
SC1 & PSPower      & 0.0182 & 0.1063 & 0.94 & 0.06 \\
SC1 & IW           & 0.0245 & 0.1206 & 0.98 & 0.02 \\
SC1 & Commensurate & 0.0187 & 0.1059 & 0.92 & 0.08 \\
SC1 & NPE          & 0.2542 & 0.3022 & 0.92 & 0.08 \\
\cmidrule(lr){1-6}
SC2 & PSPower      & 0.0039 & 0.1262 & 0.94 & 0.06 \\
SC2 & IW           & 0.0052 & 0.1422 & 0.94 & 0.06 \\
SC2 & Commensurate & 0.0087 & 0.1195 & 0.94 & 0.06 \\
SC2 & NPE          & 0.3132 & 0.3690 & 0.92 & 0.08 \\
\cmidrule(lr){1-6}
SC3 & PSPower      & 0.9591 & 0.9655 & 0.00 & 1.00 \\
SC3 & IW           & 0.6603 & 0.6707 & 0.00 & 1.00 \\
SC3 & Commensurate & 0.8593 & 0.8721 & 0.00 & 1.00 \\
SC3 & NPE          & 0.3574 & 0.3776 & 0.98 & 0.02 \\
\cmidrule(lr){1-6}
SC4 & PSPower      & 0.0132 & 0.1576 & 0.86 & 0.14 \\
SC4 & IW           & 0.0021 & 0.1551 & 0.92 & 0.08 \\
SC4 & Commensurate & 0.0053 & 0.1497 & 0.90 & 0.10 \\
SC4 & NPE          & 0.2651 & 0.3219 & 0.98 & 0.02 \\
\cmidrule(lr){1-6}
SC5 & PSPower      & 0.9551 & 0.9601 & 0.00 & 1.00 \\
SC5 & IW           & 0.6563 & 0.6687 & 0.00 & 1.00 \\
SC5 & Commensurate & 0.8571 & 0.8716 & 0.00 & 1.00 \\
SC5 & NPE          & 0.3322 & 0.3523 & 0.96 & 0.04 \\
\cmidrule(lr){1-6}
SC6 & PSPower      & 0.3229 & 0.3601 & 0.32 & 0.68 \\
SC6 & IW           & 0.0741 & 0.1675 & 0.92 & 0.08 \\
SC6 & Commensurate & 0.2082 & 0.2638 & 0.54 & 0.46 \\
SC6 & NPE          & 0.2651 & 0.3115 & 0.94 & 0.06 \\
\bottomrule
\end{longtable}
\normalsize
\renewcommand{\arraystretch}{1}

\subsection{Additional Simulation Figures}

Figure~\ref{fig:app-sim-additional} reports the remaining non-null operating
summaries. The full numerical results are given in
Tables~\ref{tab:sim-full-nonnull} and \ref{tab:sim-full-null}.

\begin{center}
\includegraphics[width=.86\textwidth]{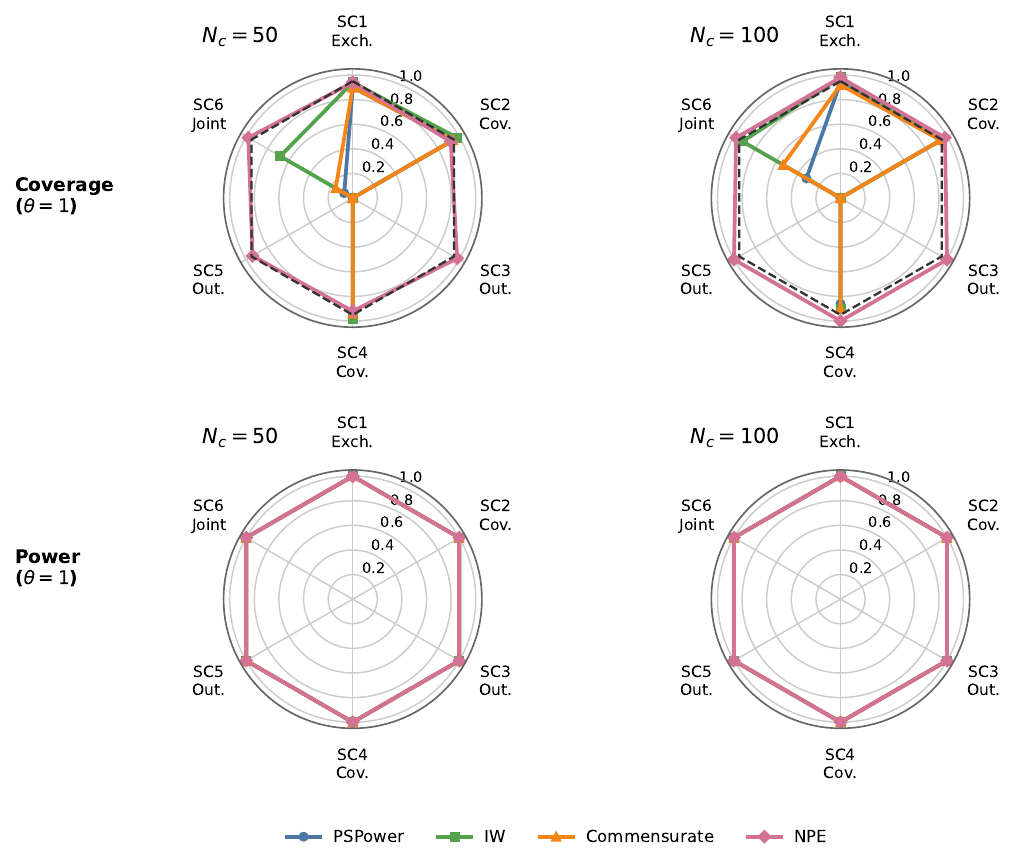}
\captionof{figure}{Additional non-null operating characteristics. Each spoke is
one simulation scenario; the dashed polygon marks nominal 95\% coverage when
applicable.}
\label{fig:app-sim-additional}
\end{center}

\begin{center}
\includegraphics[width=.76\textwidth]{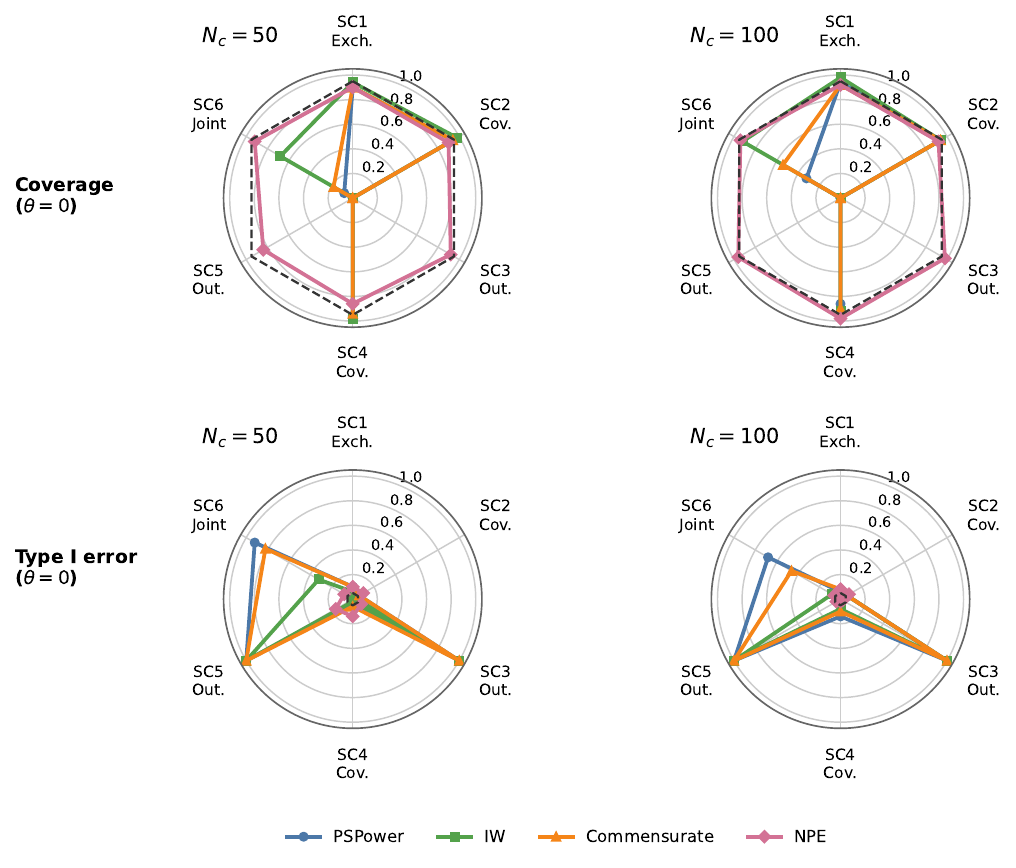}
\captionof{figure}{Null operating characteristics by scenario. Dashed polygons
indicate the nominal benchmark for each metric: 0.95 for coverage and 0.05 for
Type I error.}
\label{fig:app-sim-null-operating}
\end{center}
\FloatBarrier

\subsection{Implementation and Training Details}
\label{app:implementation}

The NPE estimator used in the main experiments is a fully connected
multilayer perceptron applied to the concatenated summary vector
$s(D_c,D_e)$. The network has hidden layers of sizes
$[128,128,64]$, ReLU activations, dropout during training, and a Gaussian
output head returning the posterior mean and log-standard-deviation of
$\theta$. Dropout is disabled at inference, so posterior evaluation for a
new current--external pair is a single deterministic forward pass after
computing the same summaries used during pretraining.
We also tested richer scalar-output density heads, including mixture-style
alternatives, during method development. They did not improve calibration or
operating characteristics for this scalar target in our experiments, so we
kept the Gaussian head for stability and simplicity.

For each simulated training triplet, the simulator draws the scalar target
$\theta$, regression coefficients $\beta$, noise scale $\sigma$, concurrent
covariate location $\mu_c$, the covariate-shift magnitude, the outcome-drift
magnitude $\delta_e$, and the mismatch type from the corresponding
pretraining distributions. The external sample size is fixed at
$n_e=200$, while the concurrent sample size is sampled from
$n_c\in\{50,100,200,500\}$ during pretraining. The evaluation grid in the
main simulation is narrower, using $n_c\in\{50,100\}$, so that the behavior
of each method can be compared under the interpretable SC1--SC6 regimes.

The conditional-structure summary block is data-derived. In the synthetic
experiments, the network is not given the true $\beta$. Instead, we fit
separate OLS models in the concurrent and external datasets and provide the
estimated coefficient differences as part of $s(D_c,D_e)$. The true
$\beta$ is used only by the simulator to generate data and to evaluate
operating characteristics.

\subsection{Perturbing the OLS-Derived Summary Block}
\label{app:ols-perturbation}

To check whether the method relies too heavily on the OLS-derived
coefficient-difference block, we reran the SC5 outcome-drift setting with
$n_c=100$ and added independent Gaussian noise to that block before
standardizing the summary vector and applying the pretrained network.
Table~\ref{tab:ols-perturbation} reports the resulting operating
characteristics over 50 simulation replicates. Coverage remains at or above
the nominal 95\% level for noise standard deviations up to 0.15, while
absolute bias increases gradually from 0.22 to 0.30. This suggests that the
coefficient-difference block is informative, but not the sole source of the
network's robustness under outcome drift.

\begin{table}[H]
\centering
\small
\caption{Sensitivity to perturbing the OLS-derived coefficient-difference
summary block in SC5 ($n_c=100$, $\theta=1$).}
\label{tab:ols-perturbation}
\begin{tabular}{ccccc}
\toprule
Noise SD & Bias & Absolute bias & RMSE & Coverage \\
\midrule
0.00 & 0.222 & 0.222 & 0.273 & 0.98 \\
0.02 & 0.226 & 0.226 & 0.276 & 0.98 \\
0.05 & 0.241 & 0.241 & 0.290 & 0.98 \\
0.10 & 0.270 & 0.270 & 0.328 & 0.98 \\
0.15 & 0.296 & 0.296 & 0.357 & 0.96 \\
\bottomrule
\end{tabular}
\end{table}

\subsection{Illustrative Covariate-Effect Target Check}
\label{app:covariate-effect}

The main empirical validation focuses on scalar current-study targets: an
additive study-effect parameter in the Gaussian simulations and a
cohort-level risk parameter in ADNI. As a limited boundary check, we also
considered a simple covariate-effect target, $\beta_1$, in the Gaussian
simulation. Because the target is now a regression coefficient rather than an
intercept-like study-effect parameter, we augmented the summary vector with
source-specific OLS slope estimates and their standard errors. We then trained
an NPE model to approximate the posterior for $\beta_1$ and applied a
validation-based scale calibration to its posterior standard deviation.

Table~\ref{tab:covariate-effect} reports results for $n_c=100$ and
$n_e=200$ over 50 simulation replicates. This check is not intended as a full
validation of arbitrary covariate effects or treatment--covariate
interactions. Rather, it shows that the same amortized framework can be
adapted to a different scalar target when the summary representation is made
target-aware. A systematic study of covariate-effect targets and their
overlap requirements remains future work.

\begin{table}[H]
\centering
\small
\caption{Limited illustrative check for a current-study covariate-effect
target, $\beta_1$ ($n_c=100$, $n_e=200$). NPE uses target-aware summaries and
validation-based posterior-scale calibration.}
\label{tab:covariate-effect}
\begin{tabular}{llcccc}
\toprule
Scenario & Method & Bias & RMSE & Coverage & Width \\
\midrule
SC1 & Concurrent OLS & -0.006 & 0.134 & 0.94 & 0.538 \\
SC1 & NPE-$\beta_1$ & 0.002 & 0.082 & 1.00 & 0.442 \\
\cmidrule(lr){1-6}
SC2 & Concurrent OLS & 0.003 & 0.136 & 0.92 & 0.542 \\
SC2 & NPE-$\beta_1$ & 0.016 & 0.064 & 1.00 & 0.459 \\
\cmidrule(lr){1-6}
SC3 & Concurrent OLS & -0.002 & 0.142 & 0.94 & 0.530 \\
SC3 & NPE-$\beta_1$ & 0.004 & 0.069 & 0.98 & 0.434 \\
\cmidrule(lr){1-6}
SC4 & Concurrent OLS & 0.002 & 0.127 & 0.94 & 0.523 \\
SC4 & NPE-$\beta_1$ & 0.021 & 0.081 & 0.98 & 0.451 \\
\cmidrule(lr){1-6}
SC5 & Concurrent OLS & -0.014 & 0.157 & 0.92 & 0.532 \\
SC5 & NPE-$\beta_1$ & -0.013 & 0.079 & 1.00 & 0.420 \\
\cmidrule(lr){1-6}
SC6 & Concurrent OLS & 0.004 & 0.151 & 0.94 & 0.548 \\
SC6 & NPE-$\beta_1$ & -0.011 & 0.073 & 0.98 & 0.422 \\
\bottomrule
\end{tabular}
\end{table}

\subsection{ADNI Prior-Scale Sensitivity}
\label{app:adni-sensitivity}

For the ADNI application, the external--concurrent outcome shift parameter
$\delta$ is a nuisance quantity used during simulation-based pretraining.
Table~\ref{tab:adni-delta-sensitivity} varies the prior standard deviation
for $\delta$ while keeping the same observed ADNI covariates and outcome
definition. Each row averages five independent NPE pretraining runs. The
posterior mean of the concurrent risk-level parameter remains close to the
empirical concurrent logit risk, $-1.584$, across the prior scales considered;
interval width changes only modestly.

\begin{table}[H]
\centering
\small
\caption{ADNI sensitivity to the prior standard deviation for the
external--concurrent outcome-shift parameter $\delta$.}
\label{tab:adni-delta-sensitivity}
\begin{tabular}{ccccc}
\toprule
$\mathrm{sd}(\delta)$ & Posterior mean & Mean SD & 95\% interval & Width \\
\midrule
0.50 & -1.573 & 0.864 & $(-3.281,\ 0.122)$ & 3.403 \\
0.75 & -1.583 & 0.862 & $(-3.287,\ 0.107)$ & 3.394 \\
1.00 & -1.587 & 0.856 & $(-3.279,\ 0.092)$ & 3.371 \\
1.25 & -1.592 & 0.854 & $(-3.279,\ 0.082)$ & 3.360 \\
\bottomrule
\end{tabular}
\end{table}

\subsection{Code Availability}
\label{app:code-availability}

The supplementary material includes runnable code for the main simulation
study, timing experiment, exploratory NPE extensions, and ADNI analysis
pipeline. Code is also available at
\url{https://github.com/ChinHungScott/NPE-for-Bayesian-Dynamic-Borrowing-MLHC-}.
The ADNI raw data are not redistributed because access requires a data-use
agreement, but the code documents the preprocessing and analysis steps needed
to reproduce the reported ADNI results once authorized data are available.

\subsection{Exploratory NPE Extensions}

We also ran exploratory non-null simulations for several NPE variants that add
explicit source-discrepancy summaries to the input representation.
The goal of these variants was to see whether additional discrepancy features
could reduce bias in the harder mismatch settings.
They also serve as a check on whether the base summary vector omits an
important source-mismatch signal for the scalar target studied here.
The added summaries include Kullback--Leibler (KL) divergence, optimal
transport (OT) distance, the Kolmogorov--Smirnov (KS) statistic, a combined
KL+KS representation, and Jensen--Shannon divergence (JSD).
These variants are not part of the main comparison, but they provide a check on
whether explicit discrepancy features change behavior in the harder simulation
settings.
Because the extensions are intended to address source mismatch, we summarize
their behavior in SC3 and SC5, which introduce outcome drift, and SC6, which
combines covariate shift and outcome drift.
Table~\ref{tab:npe-extensions-hard} reports representative results for
$N_c=100$ in these three scenarios.

\begin{table}[H]
\centering
\small
\caption{Exploratory NPE extensions in the harder non-null settings
($N_c=100$, $\theta=1$).}
\label{tab:npe-extensions-hard}
\begin{tabular}{llccc}
\toprule
Scenario & Method & Bias & RMSE & Power \\
\midrule
SC3 & NPE        & 0.1320  & 0.2223 & 1.00 \\
SC3 & KL-NPE     & 0.0242  & 0.1905 & 0.98 \\
SC3 & OT-NPE     & 0.0757  & 0.2013 & 1.00 \\
SC3 & KL+KS-NPE  & 0.0964  & 0.2037 & 1.00 \\
SC3 & JSD-NPE    & 0.1769  & 0.2498 & 1.00 \\
\cmidrule(lr){1-5}
SC5 & NPE        & 0.1483  & 0.2205 & 1.00 \\
SC5 & KL-NPE     & 0.0353  & 0.1644 & 1.00 \\
SC5 & OT-NPE     & 0.1320  & 0.3510 & 1.00 \\
SC5 & KL+KS-NPE  & 0.1278  & 0.1923 & 1.00 \\
SC5 & JSD-NPE    & 0.1880  & 0.2474 & 1.00 \\
\cmidrule(lr){1-5}
SC6 & NPE        & -0.1191 & 0.1907 & 0.96 \\
SC6 & KL-NPE     & -0.0467 & 0.1560 & 1.00 \\
SC6 & OT-NPE     & -0.0865 & 0.1605 & 0.98 \\
SC6 & KL+KS-NPE  & -0.1166 & 0.1811 & 1.00 \\
SC6 & JSD-NPE    & -0.1121 & 0.1647 & 1.00 \\
\bottomrule
\end{tabular}
\end{table}